\documentclass[11pt,a4paper]{article}
\usepackage{times,latexsym}
\usepackage{url}
\usepackage[T1]{fontenc}
\usepackage{booktabs}
\usepackage{graphicx}
\usepackage{subcaption}
\usepackage{ragged2e}
\usepackage{setspace}
\usepackage{enumitem}
\usepackage{pifont}
\usepackage{adjustbox}
\usepackage{slashbox}
\usepackage[x11names,table]{xcolor}
\usepackage[acceptedWithA]{tacl2018v2}

\newcommand*{\affmark}[1][*]{\textsuperscript{#1}}
\newcommand*{\affaddr}[1]{#1}
\usepackage{xspace,mfirstuc,tabulary}

\newif\iftaclinstructions
\taclinstructionsfalse 
\iftaclinstructions
\renewcommand{\confidential}{}
\renewcommand{\anonsubtext}{(No author info supplied here, for consistency with
TACL-submission anonymization requirements)}
\newcommand{\instr}
\fi

\iftaclpubformat 

\else

\fi

\title{Adapting to the Long Tail: A Meta-Analysis of Transfer Learning Research for Language Understanding Tasks}

\author{Aakanksha Naik\affmark[1,3] \And Jill Lehman\affmark[2] \\ 
  \affaddr{\affmark[1]Language Technologies Institute, Carnegie Mellon University} \\
  \affaddr{\affmark[2]Human-Computer Interaction Institute, Carnegie Mellon University} \\
  \affaddr{\affmark[3]Rehabilitation Medicine Department, Clinical Center, National Institutes of Health} \\
  \texttt{\{anaik,jfl,cprose\}@andrew.cmu.edu} \And Carolyn Ros\'e\affmark[1,3]}

\date{}

\begin{document}
\maketitle
\begin{abstract}
Natural language understanding (NLU) has made massive progress driven by large benchmarks, but benchmarks often leave a long tail of infrequent phenomena underrepresented. We reflect on the question: \emph{have transfer learning methods sufficiently addressed the poor performance of benchmark-trained models on the long tail?} We conceptualize the long tail using macro-level dimensions (e.g., underrepresented genres, topics, etc.), and perform a qualitative meta-analysis of 100 representative papers on transfer learning research for NLU. Our analysis asks three questions: (i) Which long tail dimensions do transfer learning studies target? (ii) Which properties of adaptation methods help improve performance on the long tail? (iii) Which methodological gaps have greatest negative impact on long tail performance? Our answers highlight major avenues for future research in transfer learning for the long tail. Lastly, using our meta-analysis framework, we perform a case study comparing the performance of various adaptation methods on clinical narratives, which provides interesting insights that may enable us to make progress along these future avenues.  
\end{abstract}

\section{Introduction}
\emph{``There is a growing consensus that significant, rapid progress can be made in both text understanding and spoken language understanding by investigating those phenomena that occur most centrally in naturally occurring unconstrained materials and by attempting to automatically extract information about language from very large corpora.''} \cite{marcus-etal-1993-building}

Since the creation of the Penn Treebank, using shared benchmarks to measure and drive progress in model development has been instrumental for accumulation of knowledge in the field of natural language processing, and has become a dominant practice. Ideally, we would like shared benchmark corpora to be diverse and comprehensive, which can be addressed at two levels: (i) macro-level dimensions such as language, genre, topic, etc., and (ii) micro-level dimensions such as specific language phenomena like negation, deixis, causal reasoning, etc. However, diversity and comprehensiveness is not straightforward to achieve. 

According to Zipf's law, many micro-level language phenomena naturally occur infrequently and will be relegated to the \emph{long tail}, except in cases of intentional over-sampling. Moreover, the advantages of restricting community focus to a specific set of benchmark corpora and limitations in resources lead to portions of the macro-level space being under-explored, which can further cause certain micro-level phenomena to be under-represented. For example, since most popular coreference benchmarks focus on English narratives, they do not contain many instances of zero anaphora, a phenomenon quite common in other languages (e.g., Japanese, Chinese). In such situations, model performance on benchmark corpora may not be truly reflective of expected performance on micro-level long tail phenomena, raising questions about the ability of state-of-the-art models to generalize to the long tail. 

Most benchmarks do not explicitly catalogue the list of micro-level language phenomena that are included or excluded in the sample, which makes it non-trivial to construct a list of long tail micro-level language phenomena. Hence, we formalize an alternate conceptualization of the long tail: undersampled portions of the macro-level space that can be treated as proxies for long tail micro-level phenomena. These undersampled \emph{long tail} macro-level dimensions highlight gaps and present potential new challenging directions for the field. Therefore, periodically taking stock of research to identify long tail macro-level dimensions can help in highlighting opportunities for progress that have not yet been tackled. This idea has been gaining prominence recently; for example, \newcite{joshi-etal-2020-state} survey languages studied by NLP papers, providing statistical support for the existence of a macro-level long tail of low-resource languages.


In this work, our goal is to attempt to characterize the macro-level long tail in NLU and efforts that have tried to address it from research on transfer learning. Large benchmarks have driven much of the recent methodological progress on NLU \cite{bowman-etal-2015-large,rajpurkar-etal-2016-squad,mccann2018natural, talmor-etal-2019-commonsenseqa,wang2019glue,wang2019superglue}, but the generalization abilities of benchmark-trained models to the long tail have been unclear. In tandem, the NLP community has been successfully developing transfer learning methods to improve generalization of models trained on NLU benchmarks \cite{ruder-etal-2019-transfer}. The goal of transfer learning research is to tackle the macro-level long tail in NLU, leading to the question: \emph{how far has transfer learning addressed performance of benchmark models on the NLU long tail, and where do we still fall behind?} 

Probing further, we perform a qualitative meta-analysis of a representative sample of 100 papers on domain adaptation and transfer learning in NLU. We sample these papers based on citation counts and publication venues (\S\ref{ssec:sample}), and document 7 facets for each paper such as tasks and domains studied, adaptation settings evaluated, etc. (\S\ref{ssec:facets}). Adaptation methods proposed (or applied) are documented using a hierarchical categorization described in~\S\ref{ssec:methodcat}, which we develop by extending the hierarchy from \newcite{ramponi-plank-2020-neural}. With this information, our analysis focuses on three questions:
\begin{itemize}[leftmargin=*,topsep=0pt]
\setlength\itemsep{-0.5em}
    \item \textbf{Q1:} What long tail macro-level dimensions do transfer learning studies target? Dimensions include tasks, domains, languages and adaptation settings covered in transfer learning research.  
    \item \textbf{Q2:} Which properties of adaptation methods help improve performance on long tail dimensions?
    \item \textbf{Q3:} Which methodological gaps have greatest negative impact on long tail performance?
\end{itemize}
The rest of the paper presents thorough answers to these questions, laying out avenues for future research on transfer learning that more effectively address the macro-level long tail in NLU. We also present a case study\footnote{The codebase for our case study experiment is available at: \href{https://github.com/CC-RMD-EpiBio/Domain-Adaptation-Meta-Analysis}{https://github.com/CC-RMD-EpiBio/Domain-Adaptation-Meta-Analysis}.} to demonstrate how our meta-analysis framework can be use to systematically design experiments that provide insights that enable us to make progress along these avenues.

\section{Meta-Analysis Framework}
\subsection{Sample Curation}
\label{ssec:sample}
\begin{figure}
    \centering
    \includegraphics[scale=0.55]{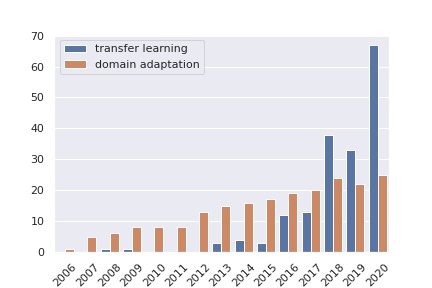}
    \caption{Distribution of papers retrieved by our search strategy across search terms and years.}
    \label{fig:keyword}
\end{figure}

We gather a representative sample of work on domain adaptation or transfer learning in NLU from the December 2020 dump of the Semantic Scholar Open Research Corpus (S2ORC) \cite{lo-etal-2020-s2orc}. First, we extract all abstracts published at 9 prestigious *CL venues: ACL, EMNLP, NAACL, EACL, COLING, CoNLL, SemEval, TACL, and CL. This results in 25,141 abstracts, which are filtered to retain those containing the terms ``domain adaptation'' or ``transfer learning'' in the title or abstract\footnote{Search scope is limited to title and abstract in order to prefer papers that focus on transfer learning rather than ones including a brief discussion or experiment on transfer learning as part of an investigation of something else.}, producing a set of 382 abstracts after duplicate removal. Figure~\ref{fig:keyword} shows the distribution of these retrieved abstracts across search terms and years. From this graph we can see that interest in this field has increased tremendously in recent years, and that there has been a slight terminology shift with recent work preferring the term ``transfer learning'' over ``domain adaptation''. 

\begin{figure}
    \centering
    \includegraphics[scale=0.32]{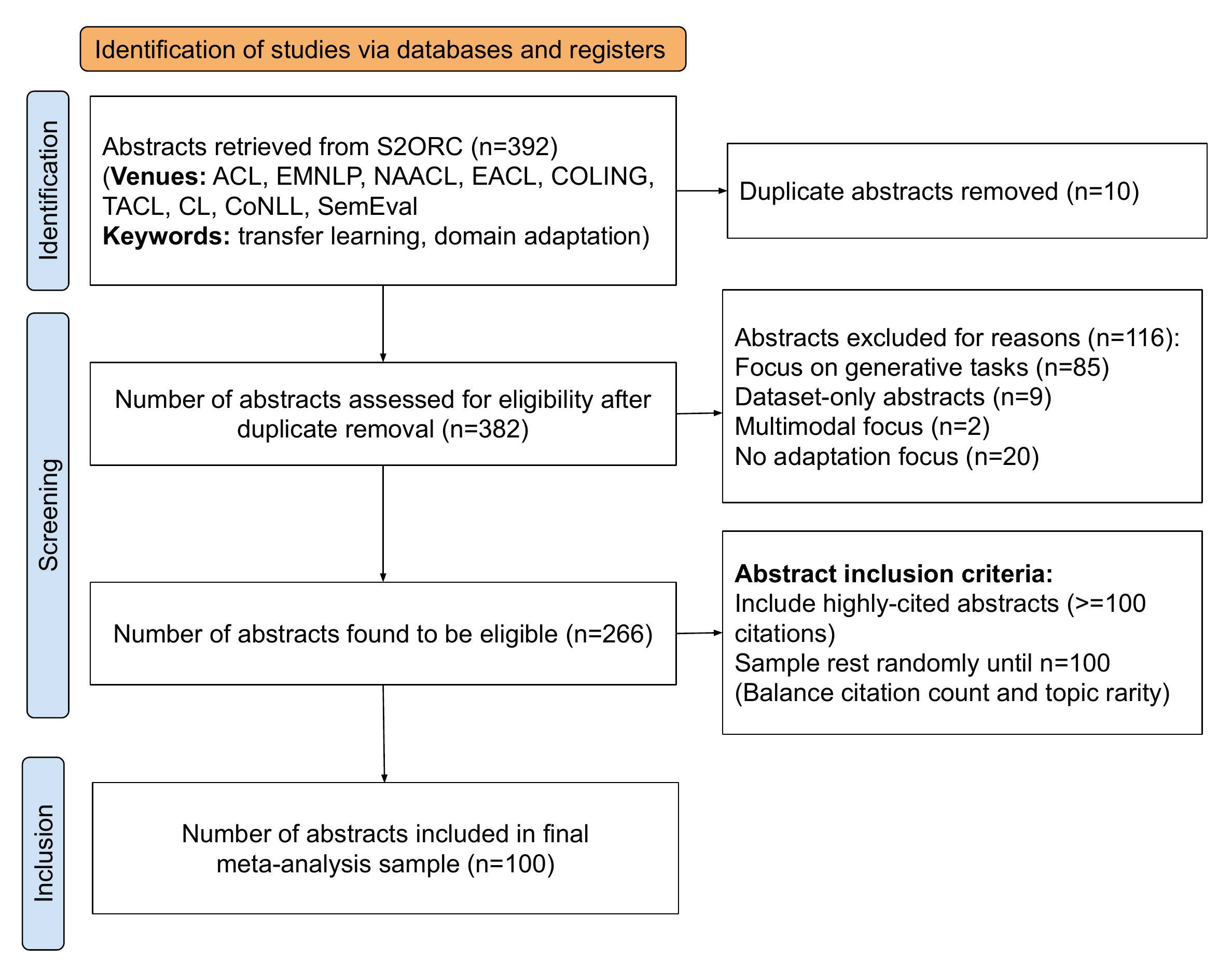}
    \caption{PRISMA diagram explaining our sample curation process}
    \label{fig:prisma}
\end{figure}

We manually screen this subset and remove abstracts that are not eligible for our NLU-focused analysis (e.g., papers on generation-focused tasks like machine translation), leaving us with a set of 266 abstracts. From this, we construct a final meta-analysis sample of 100 abstracts via application of two inclusion criteria. Per the first criterion, all abstracts with 100 or more citations are included since they are likely to describe landmark advances.\footnote{This makes up 23\% of the final meta-analysis sample.} Then, remaining abstracts (to bring our meta-analysis sample to 100) are randomly chosen, after discarding ones with no citations.\footnote{Mean citation count for randomly sampled set is 28.4.} The random sampling criterion ensures that we do not neglect studies that study less mainstream topics by focusing solely on highly-cited work. This produces a final representative sample of transfer learning work for our meta-analysis. Figure~\ref{fig:prisma} describes our sample curation process via a PRISMA (Preferred Reporting Items for Systematic Reviews and Meta-Analyses) diagram \cite{page2021prisma}.  
\begin{figure}
    \centering
    \includegraphics[scale=0.5]{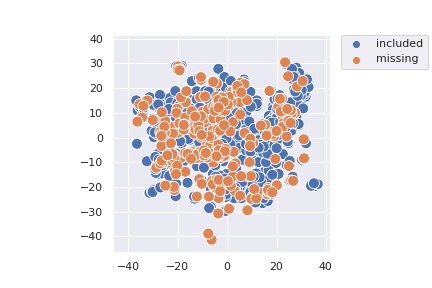}
    \caption{TSNE visualization of our meta-analysis sample alongside additional transfer learning papers missed by our keyword search. }
    \label{fig:tsne}
\end{figure}

\begin{table}[]
\footnotesize
    \centering
    \begin{tabular}{lp{5.5cm}}
    \toprule \textbf{Cat} & \textbf{Tasks Included} \\ \midrule
    TC & Text classification tasks like sentiment analysis, hate speech detection, propaganda detection, etc.\\
    NER & Semantic sequence labeling tasks like NER, event extraction, etc.\\
    POS & Syntactic sequence labeling tasks like POS tagging, chunking, etc.\\
    NLI & Natural language inference, NLU Tasks recast as NLI (e.g., GLUE)\\
    SP & Structured prediction tasks such as entity and event coreference\\
    WSD & Word sense disambiguation \\
    TRN & Text ranking tasks (e.g., search)\\
    TRG & Text regression tasks\\
    RC & Reading comprehension\\
    MF & Matrix factorization\\
    LI & Lexicon induction\\
    SLU & Spoken language understanding\\ \bottomrule
    \end{tabular}
    \caption{Categorization of tasks studied. Note that the matrix factorization category includes text-based recommender systems.}
    \label{tab:taskcat}
    \vspace{-4mm}
\end{table}

\noindent
\textbf{Characterizing limitations of our curation process:} Since our sample curation process primarily relies on a keyword-based search, it might miss relevant work that does not use any of these keywords. To characterize the limitations of our curation process, we employ two additional strategies for relevant literature identification:
\begin{itemize}[leftmargin=*,topsep=0pt]
\setlength\itemsep{-0.5em}
    \item \textbf{Citation graph retrieval:} Following \citet{blodgett-etal-2020-language}, we include all abstracts that cite or are cited by abstracts included in our keyword-retrieved set of 382 abstracts. This retrieves 3727 additional abstracts, but many of these works are cited for their description or introduction of new tasks, datasets, evaluation metrics, etc. Therefore, we discard all works that do not have the words ``adaptation'' or ``transfer'', leaving 282 new abstracts. 
    \item \textbf{Nearest neighbor retrieval:} We use SPECTER \citep{cohan-etal-2020-specter} to compute embeddings for all abstracts included in our keyword-retrieved set, as well as all abstracts in the ACL anthology. Then we retrieve the nearest neighbor for every abstract in our keyword-retrieved set, which results in the retrieval of 262 new abstracts.
\end{itemize}

\begin{figure*}[]
    \centering
    \includegraphics[scale=0.23]{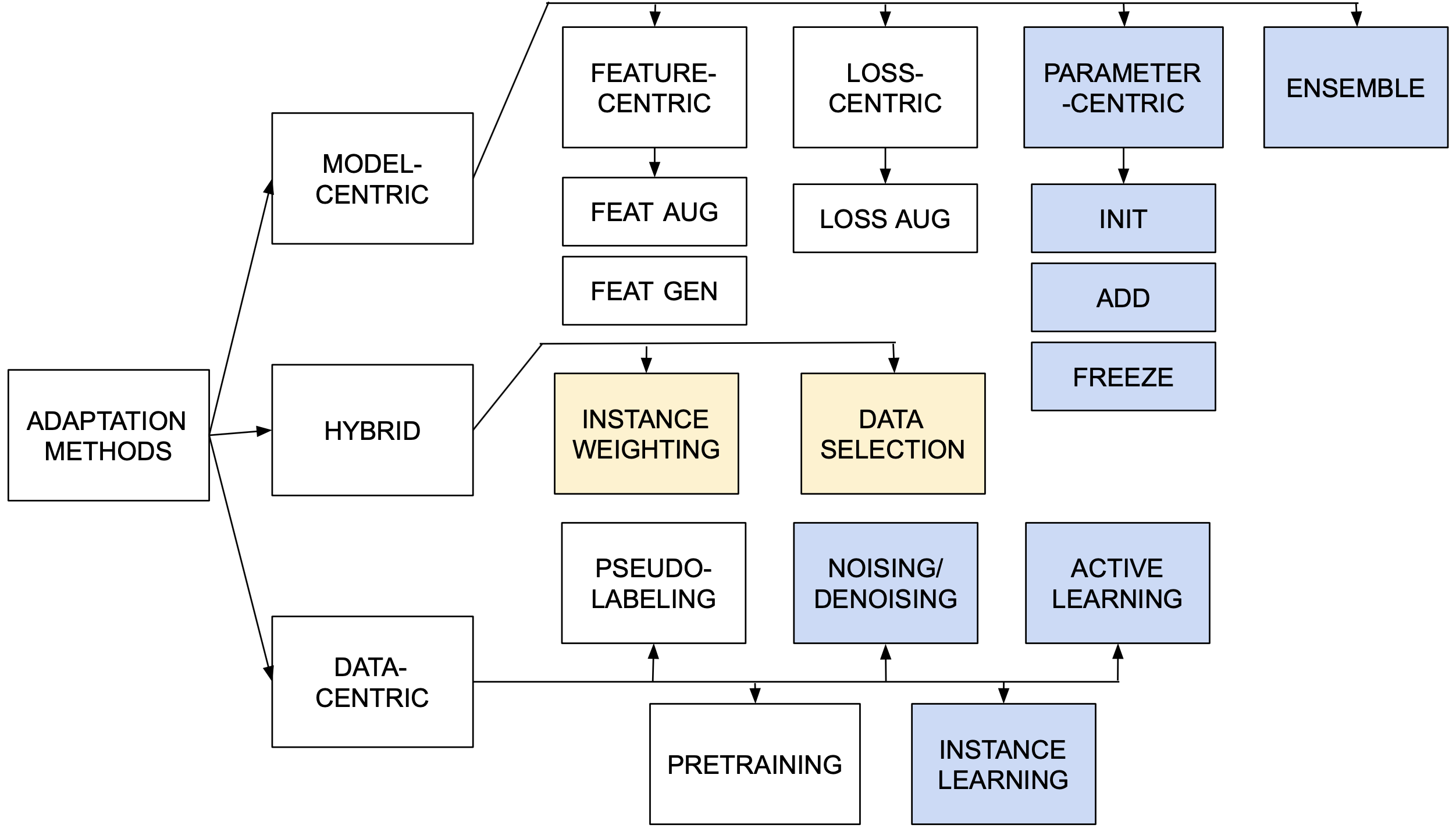}
    \caption{Categorization of adaptation methods proposed, extended or used in all studies. This categorization is an extension of the one proposed by \citet{ramponi-plank-2020-neural}, with blue blocks indicating newly added categories, and yellow blocks indicating categories that have been moved to a different coarse category.}
    \label{fig:methodcat}
\end{figure*}

Combining abstracts returned by both strategies, we are able to identify 510 additional works. However, while going over them manually, we notice that despite our noise reduction efforts, not all abstracts describe transfer learning work. We perform an additional manual screening step to discard such work, which leaves us with a final set of 232 additional papers.\footnote{We make this subset of papers available at: \href{http://www.shorturl.at/uFGIY}{http://www.shorturl.at/uFGIY}} To identify whether the exclusion of these papers from the initial sample may have led to visible gaps or blind spots in our meta-analysis, we perform a TSNE visualization of SPECTER embeddings for both keyword-retrieved papers and this additional set of papers. Figure~\ref{fig:tsne} presents the results of this visualization and indicates that there aren't visible distributional differences between the two subsets. Hence, though our sample curation strategy is imperfect, this seems to indicate that our final observations from the meta-analysis might not have been very different. We note that this conclusion comes with two caveats: (i) t-SNE embeddings are not always reliable, and (ii) embedding overlap does not necessarily confirm that annotations for overlapping papers are similar/correlated. Keeping these caveats in mind, we perform a spot check for additional validation. For this spot check, we consider the following highly-cited large language models that have been considered to be major recent advances in transfer learning: ELMo, BERT, RoBERTa, BART, T5, ERNIE, DeBERTa, and ELECTRA. Note that we do not consider any few-shot models (e.g., GPT3, PET, etc.) since our sample only consists of work that was \emph{accepted} to a *CL venue by December 2020. Of these major language models, RoBERTa, DeBERTa, T5, and ELECTRA were published at non-*CL venues (JMLR and ICLR), which excludes them from our sample. The remaining works (ELMo, BERT, BART, and ERNIE) are all present in the set of additional works we identified in this section, lending further support to our conclusion that our sampling strategy and subsequent analyses have not overlooked influential work. 


\subsection{Meta-Analysis Facets}
\label{ssec:facets}
For every paper from our meta-analysis sample, we document the following key facets:\\
\noindent
\textbf{Task(s):} NLP task(s) studied in the work. Tasks are grouped into 12 categories based on task formalization and linguistic level (e.g., lexical, syntactic, etc.), as shown in table~\ref{tab:taskcat}.\\
\noindent
\textbf{Domain(s):} Source and target domains and/or languages studied, along with datasets used for each. \\
\noindent
\textbf{Task Model:} Base model used for the task, to which domain adaptation algorithms are applied.\\
\noindent
\textbf{Adaptation Method(s):} Domain adaptation method(s) proposed or used in the work. Adaptation methods are grouped according to the categorization showed in figure~\ref{fig:methodcat} (details in~\S\ref{ssec:methodcat}). \\
\noindent
\textbf{Adaptation Baseline(s):} Baseline domain adaptation method(s) to compare new methods against.\\
\noindent
\textbf{Adaptation Settings:} Source-target transfer settings explored in the work (e.g., unsupervised adaptation, multi-source adaptation, etc.).\\
\noindent
\textbf{Result Summary:} Performance improvements (if any), performance differences across multiple source-target pairs or methods, etc.
\subsection{Adaptation Method Categorization}
\label{ssec:methodcat}
For adaptation methods proposed or used in each study, we assign type labels according to the categorization presented in figure~\ref{fig:methodcat}. This categorization is an extension of the one proposed by \newcite{ramponi-plank-2020-neural}.\footnote{Since our meta-analysis is not limited to neural unsupervised domain adaptation, we need to add additional classes.} Broadly, methods are divided into three \emph{coarse} categories: (i) model-centric, (ii) data-centric, and (iii) hybrid approaches. Model-centric approaches perform adaptation by modifying the structure of the model, which may include editing the feature representation, loss function or parameters. Data-centric approaches perform adaptation by modifying or leveraging labeled/unlabeled data from the source and target domains to bridge the domain gap. Finally, hybrid approaches are ones that cannot be clearly classified as model-centric or data-centric. Each coarse category is divided into \emph{fine} subcategories.
\begin{table*}[ht]
    \centering
    \footnotesize
    \begin{tabular}{>{\raggedright}p{3.5cm}p{6.25cm}p{5.0cm}}
    \toprule \textbf{Category} & \textbf{Example Methods} & \textbf{Example Studies} \\ \midrule
    Feat Aug (FA) & Structural correspondence learning, Frustratingly easy domain adaptation &  \cite{blitzer-etal-2006-domain,daume-iii-2007-frustratingly}\\ \midrule
    Feat Gen (FG) & Marginalized stacked denoising autoencoders, Deep belief networks & \cite{jochim-schutze-2014-improving,ji-etal-2015-closing,yang-etal-2015-domain} \\ \midrule
    Loss Aug (LA) & Multi-task learning, Adversarial learning, Regularization-based methods & \cite{zhang-etal-2017-aspect,liu-etal-2019-multi,chen-etal-2020-recall}\\ \midrule
    Init (PI) & Prior estimation, Parameter matrix initialization & \cite{chan-ng-2006-estimating,al-boni-etal-2015-model} \\ \midrule
    Add (PA) & Adapter networks & \cite{lin-lu-2018-neural} \\ \midrule
    Freeze (FR) & Embedding freezing, Layerwise freezing & \cite{yin-etal-2015-online,tourille-etal-2017-limsi} \\ \midrule
    Ensemble (EN) & Mixture of experts, Weighted averaging & \cite{mcclosky-etal-2010-automatic,nguyen-etal-2014-robust}\\ \midrule
    Instance Weighting (IW) & Classifier based weighting & \cite{jiang-zhai-2007-instance,jeong-etal-2009-semi}\\ \midrule
    Data Selection (DS) & Confidence-based sample selection & \cite{scheible-schutze-2013-sentiment,braud-denis-2014-combining} \\ \midrule
    Pseudo-Labeling (PL) & Semi-supervised learning, Self-training & \cite{umansky-pesin-etal-2010-multi,lison-etal-2020-named}\\ \midrule
    Noising/Denoising (NO) & Token dropout & \cite{pilan-etal-2016-predicting}\\ \midrule
    Active Learning (AL) & Sample selection via active learning & \cite{rai-etal-2010-domain,wu-etal-2017-active} \\ \midrule
    Pretraining (PT) & Language model pretraining, Supervised pretraining & \cite{conneau-etal-2017-supervised,howard-ruder-2018-universal}\\ \midrule
    Instance Learning (IL) & Nearest neighbor learning & \cite{gong-etal-2016-modeling} \\ \bottomrule
    \end{tabular}
    \caption{Examples of methods from each category, and papers studying these methods. These lists are non-exhaustive. In the interest of replicability, we have made our coding for all papers publicly available at: \href{http://www.shorturl.at/stuAT}{http://www.shorturl.at/stuAT}.}
    \label{tab:methodex}
    \vspace{-4.5mm}
\end{table*}

Model-centric approaches are divided into four categories, based on which portion of the model they modify: (i) feature-centric, (ii) loss-centric, (iii) parameter-centric, and (iv) ensemble. Feature-centric approaches are further divided into two fine subcategories: (i) feature augmentation, and (ii) feature generalization. Feature augmentation includes techniques that learn an alignment between source and target feature spaces using shared features called \emph{pivots} \cite{blitzer-etal-2006-domain}. Feature generalization includes methods that learn a joint representation space using autoencoders, motivated by \newcite{glorot2011domain,chen2012marginalized}. Loss-centric approaches contain one fine subcategory: loss augmentation. This includes techniques which augment task loss with adversarial loss \cite{ganin2015unsupervised,ganin2016domain}, multi-task loss \cite{liu-etal-2019-multi} or regularization terms. Parameter-centric approaches include three fine subcategories: (i) parameter initialization, (ii) new parameter addition, and (iii) parameter freezing. Finally ensemble, used in settings with multiple source domains, includes techniques that learn to combine predictions of multiple models trained on source and target domains.

Data-centric approaches are divided into five fine subcategories. Pseudo-labeling approaches train classifiers that then produce ``gold'' labels for unlabeled target data. This includes semi-supervised learning methods such as bootstrapping, co-training, self-training, etc. (e.g., \newcite{mcclosky-etal-2006-effective}). Active learning approaches use a human-in-the-loop setting to annotate a subset of target data that the model can learn most from \cite{settles2009active}. Instance learning approaches leverage neighborhood structure in joint source-target feature spaces to make target predictions (e.g., nearest neighbor learning). Noising/denoising approaches include data corruption/pre-processing which increase surface similarity between source and target examples. Finally, pretraining includes approaches that train large-scale language models on unlabeled data to learn better source and target representations, a strategy that has gained popularity in recent years \cite{gururangan-etal-2020-dont}.

\begin{figure}
     \centering
     \includegraphics[scale=0.4]{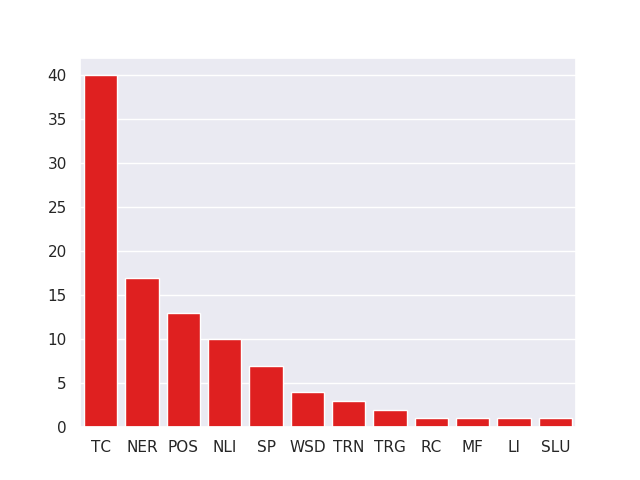}
     \caption{Distribution of papers according to tasks studied. Top three task categories are text classification (TC), semantic sequence labeling (NER) and syntactic sequence labeling (POS). Table~\ref{tab:taskcat} describes the remaining task categories.}
     \label{fig:task}
\vspace{-5mm}
 \end{figure}

Hybrid approaches contain two fine subcategories that cannot be classified as model-centric or data-centric because they involve manipulation of the data distribution, but can also be viewed as loss-centric approaches that edit the training loss. Instance weighting approaches assign weights to target examples based on similarity to source data. Conversely, data selection approaches filter target data based on similarity to source data. Table~\ref{tab:methodex} lists example adaptation methods for each fine category and example studies from our meta-analysis subset that use these methods. 

\begin{figure}
     \centering
     \includegraphics[scale=0.4]{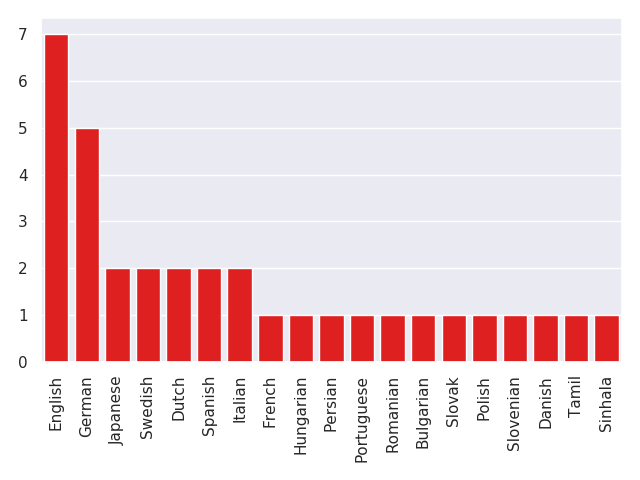}
     \caption{Distribution of multi-lingual studies according to languages included.}
     \label{fig:lang}
\end{figure} 

\section{Which Long Tail Macro-Level Dimensions Do Transfer Learning Studies Target?}
\label{sec:lttrends}
The first goal of our meta-analysis is to document long tail macro-level dimensions that transfer learning studies have tested their methods on. We look at distributions of tasks, domains, languages and adaptation settings studied in all papers in our sample. Ten studies are surveys, position papers or meta-experiments, and so excluded from these statistics. Studies can cover multiple tasks, domains, languages or settings so counts may be higher than 90.

\noindent
\textbf{Task distribution:} Figure~\ref{fig:task} gives a brief overview of the distribution of tasks studied across papers. Text classification tasks clearly dominate, followed by semantic and syntactic tagging. Text classification covers a variety of tasks, but sentiment analysis is the most well-studied, with research driven by the multi-domain sentiment detection (MDSD) dataset \cite{blitzer-etal-2007-biographies}. Conversely, structured prediction is under-studied ($<$10\% studies from our sample evaluate on structured prediction tasks), despite covering a variety of tasks such as coreference resolution, syntactic parsing, dependency parsing, semantic parsing, etc. This indicates that tasks with complex formulations/objectives are under-explored. We speculate that there may be two reasons for this: (i) difficulty of collecting annotated data in multiple domains/languages for such tasks,\footnote{Note that despite these difficulties, efforts to collect data for structured prediction tasks are underway, such as the massive Universal Dependencies project which has collected consistent grammar annotations for over 100 languages: \href{https://universaldependencies.org}{https://universaldependencies.org}} and (ii) shift in output structures (e.g., different named entity types in source and target domains) making adaptation harder.
 \begin{table}[]
\footnotesize
    \centering
    \begin{tabular}{lrlr}
    \toprule \textbf{HE} & \textbf{\#P} & \textbf{NN} & \textbf{\#P}  \\ \midrule
        Clinical & 10 & Twitter & 12\\
        Biomedical & 9 & Conversations & 10\\
        Science & 3 & Forums & 8\\
        Finance & 3 & Emails & 6\\
        Literature & 3 &&\\
        DefSec & 1 &&\\ \bottomrule
    \end{tabular}
    \caption{Counts of papers (\#P) studying high-expertise (HE) and non-narrative (NN) domains (DefSec refers to security and defense reports).}
    \label{tab:exdom}
\end{table}
\begin{figure}
     \centering
     \includegraphics[scale=0.4]{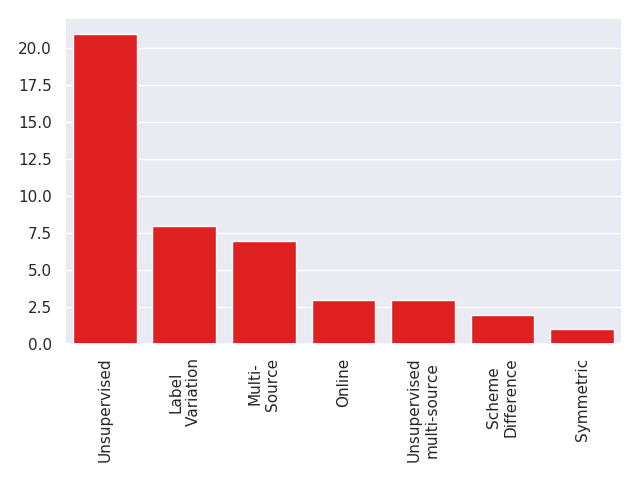}
     \caption{Distribution of papers according to unconventional (non-supervised) adaptation settings}
     \label{fig:setting}
 \end{figure}
 
\noindent
\textbf{Languages studied:} Despite a focus on generalization, most studies in our sample rarely evaluate on other languages aside from English. As stated by \newcite{bender2011achieving}, this is problematic because the ability to apply a technique to other languages does not necessarily guarantee comparable performance. Some studies do cover multi-lingual evaluation or focus on cross-linguality. Figure~\ref{fig:lang} shows the distribution of languages included in these studies, which is a limited subset. For a more comprehensive discussion of linguistic diversity in NLP research not limited to transfer learning, we refer interested readers to \newcite{joshi-etal-2020-state}.

\noindent
\textbf{Domains studied:} Many popular transfer benchmarks \cite{blitzer-etal-2007-biographies,wang2019glue,wang2019superglue} are homogeneous. They focus on narrative English, drawn from plentiful sources such as news articles, reviews, blogs, essays and Wikipedia. This sidelines some categories of domains\footnote{\emph{Domain} is an overloaded term covering genres, styles, registers, etc., but we use it for consistency with prior work.} that fall into the long tail: (i) non-narrative text (e.g., social media, conversations etc.), and (ii) texts from high-expertise domains that use specialized vocabulary and knowledge (e.g., clinical text). Table~\ref{tab:exdom} shows the number of papers focusing on high-expertise and non-narrative domains, highlighting the lack of focus on these areas.
\begin{figure}
    \centering
    \includegraphics[scale=0.35]{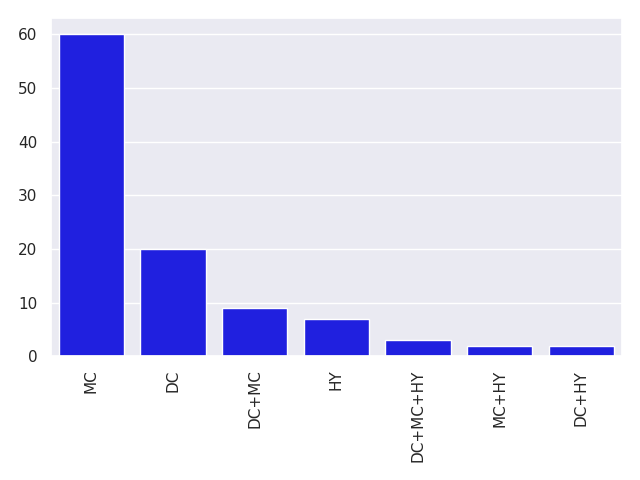}
    \caption{Distribution of transfer learning studies according to coarse method categories. DC, MC and HY refer to data-centric, model-centric, and hybrid coarse categories respectively.}
    \label{fig:coarse}
\end{figure}

\noindent
\textbf{Adaptation settings studied:} Most studies evaluate methods in a supervised adaptation setting, i.e. labeled data is available from both source and target domains. This assumption may not always hold. Often adaptation must be performed in harder settings such as unsupervised adaptation (no labeled data from target domain), adaptation from multiple source domains, online adaptation, etc, and we refer to all such settings aside from supervised adaptation as unconventional adaptation settings. Figure \ref{fig:setting} shows the distribution of unconventional settings across papers, indicating that these settings are understudied in literature.

\begin{figure}
    \centering
    \includegraphics[scale=0.45]{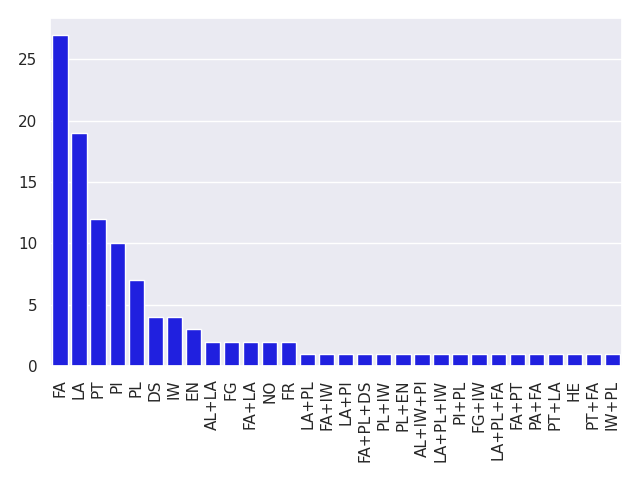}
    \caption{Distribution of transfer learning studies according to fine method categories. The top five fine categories are feature augmentation (FA), loss augmentation (LA), pretraining (PT), parameter initialization (PI), and pseudo-labeling (PL). Table~\ref{tab:methodex} describes the remaining categories in more detail.}
    \label{fig:fine}
    \vspace{-5mm}
\end{figure}
\begin{figure*}[h]
     \centering
     \begin{subfigure}{0.3\textwidth}
         \centering
         \includegraphics[width=\textwidth]{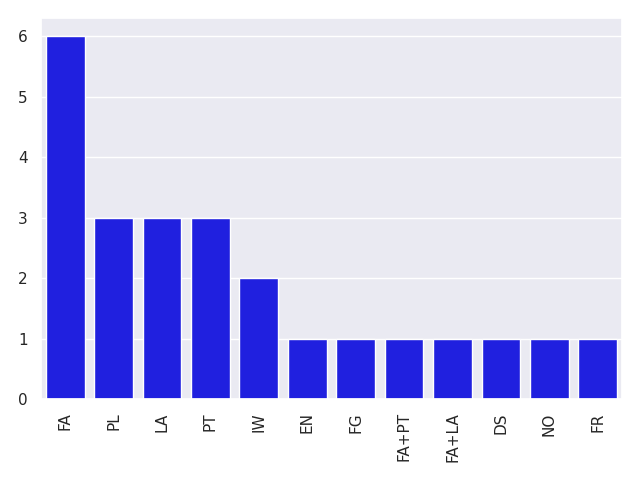}
         \caption{Fine method categories evaluated on high-expertise domains. The top five fine categories are FA, PL, LA, PT, and IW.}
         \label{fig:exfine}
     \end{subfigure}
     \hfill
     \begin{subfigure}{0.3\textwidth}
         \centering
         \includegraphics[width=0.9\textwidth]{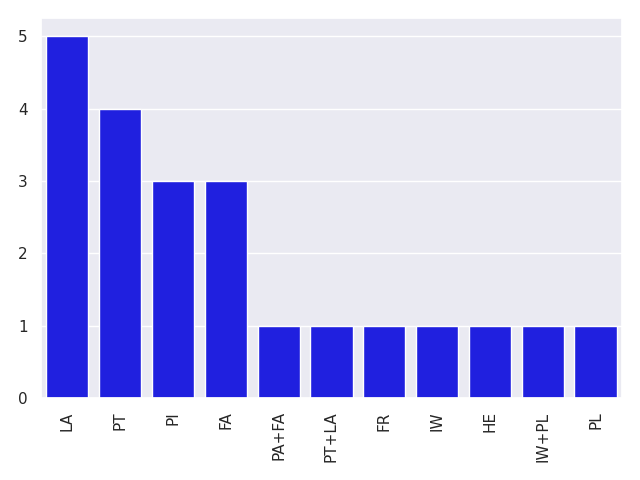}
         \caption{Fine method categories evaluated on non-narrative domains. The top four fine categories are LA, PT, PI, and FA.}
         \label{fig:nnfine}
     \end{subfigure}
     \hfill
     \begin{subfigure}{0.3\textwidth}
         \centering
         \includegraphics[width=\textwidth]{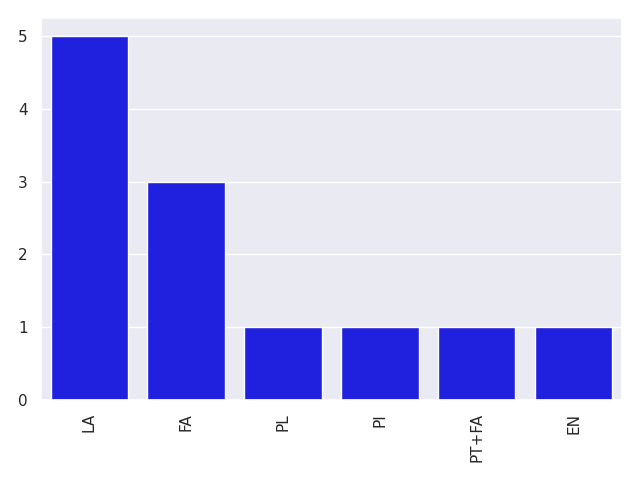}
         \caption{Fine method categories evaluated on both domain types. The top four fine categories are LA, FA, PL, and PI.}
         \label{fig:bothfine}
     \end{subfigure}
     \caption{Distribution of fine method categories from studies evaluating on long tail domains. Note that FA stands for feature augmentation, LA for loss augmentation, PL for pseudo-labeling, PT for pretraining, IW for instance weighting, and PI for parameter initialization. Table~\ref{tab:methodex} describes the remaining categories in more detail.}
\end{figure*}

\noindent
\textbf{Open Issues:} We can see that there is much ground to cover in testing adaptation methods on the macro long tail. Two research directions may be key to achieving this: (i) development of and evaluation on diverse benchmarks, and (ii) incentivizing publication of research on long tail domains at NLP venues. Diverse benchmark development has gained momentum, with the creation of benchmarks such as BLUE \cite{peng-etal-2019-transfer} and BLURB \cite{gu2020domain} for biomedical and clinical NLP, XTREME \cite{hu2020xtreme} for cross-lingual NLP and GLUECoS \cite{khanuja-etal-2020-gluecos} for code-switched NLP. However, newly proposed adaptation methods are often not evaluated on them, which is imperative to test their limitations and generalization abilities. On the other hand, application-specific or domain-specific evaluations of adaptation methods are sidelined at NLP venues and may be viewed as limited in terms of bringing broader insights. But applied research can unearth significant opportunities for advances in transfer learning, and should be viewed from a \emph{translational} perspective \cite{newman2021translational}. For example, source-free domain adaptation in which only a trained source model is available with no access to source data \cite{liang2020we}, was conceptualized partly due to data sharing restrictions on Twitter or clinical data. Though this issue is limited to certain domains, source-free adaptation may be of broader interest since it has implications for reducing models' reliance on large amounts of data. Therefore, encouraging closer ties with applied transfer learning research can help us gain more insight into limitations of existing techniques on the macro long tail.

\begin{table*}[t]
    \centering
    \footnotesize
    \begin{tabular}{lp{7cm}p{4.5cm}}
    \toprule \textbf{Study} & \textbf{Method} & \textbf{Performance} \\ \midrule
     \cite{arnold-etal-2008-exploiting} & Manually constructed feature hierarchy across domains, allowing back off to more general features (FA) & Positive transfer from 5 corpora (biomedical, news, email) to email \\
     \cite{mcclosky-etal-2010-automatic} & Mixture of domain-specific models chosen via source-target similarity features (e.g., cosine similarity) (EN) & Positive transfer to biomedical, literature and conversation domains\\
     \cite{yang-eisenstein-2015-unsupervised} & Dense embeddings induced from template features and manually defined domain attribute embeddings (FA) & Positive transfer to 4/5 web domains and 10/11 literary periods\\
     \cite{xing-etal-2018-adaptive} & Multi-task learning method with source-target distance minimization as additional loss term (LA) & Positive transfer on 4/6 intra-medical settings (EHRs, forums) and 5/9 narrative to medical settings\\
     \cite{wang-etal-2018-label-aware} & Source-target distance minimized using two loss penalties (LA) & Positive transfer to medical and Twitter data \\
    \bottomrule
    \end{tabular}
    \caption{Model and performance details for studies testing on high-expertise and non-narrative domains. Fine method categories used in these studies include feature augmentation (FA), loss augmentation (LA), ensembling (EN), pretraining (PT), parameter initialization (PI), and pseudo-labeling (PL).}
    \label{tab:longtailmethods}
    \vspace{-3mm}
\end{table*}

\section{Which Properties Of Adaptation Methods Help Improve Performance On Long Tail Dimensions?}
The second goal of our meta-analysis is to identify which categories of adaptation methods have been tested extensively and have exhibited good performance on various long tail macro-level dimensions. Figures~\ref{fig:coarse} and~\ref{fig:fine} provide an overview of categories of methods tested across all papers in our subset. We can see that studies overwhelmingly develop or use model-centric methods. Within this coarse category, feature augmentation (FA) and loss augmentation (LA) are the top two categories, followed by pretraining (PT), which is data-centric. Parameter initialization (PI) and pseudo labeling (PL) round out the top five. Feature augmentation being the most explored category is no surprise, given that a lot of pioneering early domain adaptation work in NLP \cite{blitzer-etal-2006-domain,blitzer-etal-2007-biographies,daume-iii-2007-frustratingly} developed methods to learn shared feature spaces between source and target domains. Loss augmentation methods have gained prominence recently, with multi-task learning providing large improvements \cite{liu-etal-2015-representation,liu-etal-2019-multi}. Pretraining methods, both unsupervised \cite{howard-ruder-2018-universal} and supervised \cite{conneau-etal-2017-supervised}, have also gained popularity with large transformer-based language models (e.g.,  \newcite{peters-etal-2018-deep}, \newcite{devlin-etal-2019-bert}, etc.) achieving huge gains across tasks.

To specifically identify techniques that work on long tail domains, we look at categories of methods evaluated on high-expertise domains or non-narrative domains (or both). Figures~\ref{fig:exfine},~\ref{fig:nnfine} and~\ref{fig:bothfine} present the distributions of fine method categories tested on high-expertise domains, non-narrative domains and both domain types respectively. While feature augmentation techniques remain the most explored category for high-expertise domains, we see a change in trend for non-narrative domains. Loss augmentation and pretraining are more commonly explored categories. The difference in dominant model categories can be partly attributed to easy availability of large-scale unlabeled data and weak signals (e.g., likes, shares etc.), particularly for social media. Such user-generated content (called ``fortuitous data'' by \newcite{DBLP:conf/konvens/Plank16}) is leveraged well by pretraining or multi-task learning techniques, making them popular choices for non-narrative domains. In contrast, high-expertise domains (e.g, security and defense reports, finance, etc.) often lack fortuitous data, with methods developed for them focusing on learning shared feature spaces. 

Ten studies in our meta-analysis sample evaluate on both domain types. Five of these studies (described in table~\ref{tab:longtailmethods}) operationalize two key ideas that seem to improve adaptation performance but have remained relatively under-explored in the context of recent methods like pretraining:
\begin{itemize}[leftmargin=*,topsep=0pt]
\setlength\itemsep{-0.5em}
    \item \textbf{Incorporating source-target distance:} Several methods explicitly incorporate distance between source and target domain (e.g., \cite{xing-etal-2018-adaptive,wang-etal-2018-label-aware}). Aside from allowing flexible adaptation based on the specific domain pairs being considered, adding source-target distance provides two benefits. It offers an additional avenue to analyze generalizability by monitoring source-target distance during adaptation. It also allows performance to be estimated in advance using source-target distance, which can be helpful when choosing an adaptation technique for a new target domain. \newcite{kashyap2020domain} provide a comprehensive overview of source-target distance metrics and discuss their utility in analysis and performance prediction. Despite these benefits, very little work has tried to incorporate source-target distance into newer adaptation methods such as pretraining.
    \item \textbf{Incorporating nuanced domain variation:} Despite NLP treating domain variation as a dichotomy (source vs target), domains vary along a multitude of dimensions (e.g., topic, genre, medium of communication etc.) \cite{DBLP:conf/konvens/Plank16}. Some methods acknowledge this nuance and treat domain variation as multi-dimensional, either in a discrete feature space \cite{arnold-etal-2008-exploiting} or in a continuous embedding space \cite{yang-eisenstein-2015-unsupervised}. This allows knowledge sharing across dimensions common to both source and target, improving transfer. This idea has also remained under-explored, though recent work such as the development of domain expert mixture (DEMix) layers \cite{gururangan2021demix} has attempted to incorporate nuanced domain variation into pretraining.
\end{itemize}

\noindent
\textbf{Open Issues:} Interestingly many studies from our sample do not analyze failures, i.e., source-target pairs on which adaptation methods do not improve performance. For some studies in table~\ref{tab:longtailmethods}, adaptation methods do not improve performance on all source-target pairs. But failures are not investigated, presenting the question: \emph{do we know blind spots for current adaptation methods?} Answering this is essential to develop a complete picture of the generalization capabilities of adaptation methods. Studies that present negative transfer results (e.g., \cite{plank-etal-2014-importance}) are rare, but should be encouraged to develop a sound understanding of adaptation techniques. Analyses should also study ties between datasets used and methods applied, highlighting dimensions of variation between source-target domains and how adaptation methods bridge them \cite{kashyap2020domain,naik-etal-2021-adapting}. Such analyses can uncover important lessons about generalizability of adaptation methods and the kinds of source-target settings they can be expected to improve performance on. 

\begin{table*}[]
    \centering
    \tiny
    \begin{tabular}{lcccccccccccc}
      \toprule \backslashbox{\textbf{M}}{\textbf{T}} & \textbf{TC} & \textbf{POS} & \textbf{NER} & \textbf{NLI} & \textbf{SP} & \textbf{WSD} & \textbf{TRN} & \textbf{TRG} & \textbf{RC} & \textbf{MF} & \textbf{LI} & \textbf{SLU}\\ \midrule
       \textbf{Feature Augmentation} & \cellcolor{blue!35}25 & \cellcolor{blue!10}6 & \cellcolor{blue!20}13 & \cellcolor{blue!10}3 & \cellcolor{blue!10}4 & \cellcolor{blue!10}2 & \cellcolor{blue!10}1 & \cellcolor{blue!10}2 & \cellcolor{blue!10}1 & & \cellcolor{blue!10}1 & \cellcolor{blue!10}1 \\
       \textbf{Feature Generalization} & \cellcolor{blue!10}2 & \cellcolor{blue!10}1 & \cellcolor{blue!10}1 & \cellcolor{blue!10}1 & \cellcolor{blue!10}1 & & & & & & & \\ 
       \textbf{Loss Augmentation} & \cellcolor{blue!35}21 & \cellcolor{blue!10}5 & \cellcolor{blue!10}7 & \cellcolor{blue!10}4 & \cellcolor{blue!10}3 & & 1\cellcolor{blue!10} & & \cellcolor{blue!10}2 & & \cellcolor{blue!10}1 & \\ 
       \textbf{Parameter Initialization} & \cellcolor{blue!10}7 & \cellcolor{blue!10}1 & \cellcolor{blue!10}4 & \cellcolor{blue!10}2 & & \cellcolor{blue!10}2 & \cellcolor{blue!10}1 & & & \cellcolor{blue!10}1 & & \\ 
       \textbf{Parameter Addition} & & & \cellcolor{blue!10}1 & & & & & & & & & \\ 
       \textbf{Parameter Freezing} & \cellcolor{blue!10}1 & \cellcolor{blue!10}1 & \cellcolor{blue!10}1 & & & & & & & & & \\ 
       \textbf{Ensemble} & \cellcolor{blue!10}2 & & \cellcolor{blue!10}1 & \cellcolor{blue!10}1 & \cellcolor{blue!10}2 & & & & & & & \\ 
       \textbf{Instance Weighting} & \cellcolor{blue!10}9 & \cellcolor{blue!10}3 & \cellcolor{blue!10}1 & \cellcolor{blue!10}1 & & \cellcolor{blue!10}1 & & & & & & \\ 
       \textbf{Data Selection} & \cellcolor{blue!10}3 & & \cellcolor{blue!10}1 & \cellcolor{blue!10}1 & \cellcolor{blue!10}1 & & & & & & & \\ 
       \textbf{Pretraining} & \cellcolor{blue!20}13 & & \cellcolor{blue!10}2 & \cellcolor{blue!10}9 & \cellcolor{blue!10}3 & & & \cellcolor{blue!10}1 & & & \cellcolor{blue!10}1 & \\ 
       \textbf{Pseudo-Labeling} & \cellcolor{blue!10}9 & \cellcolor{blue!10}3 & \cellcolor{blue!10}3 & \cellcolor{blue!10}1 & \cellcolor{blue!10}4 & & & & \cellcolor{blue!10}1 & & & \\ 
       \textbf{Noising/Denoising} &\cellcolor{blue!10} 2 & \cellcolor{blue!10}1 & \cellcolor{blue!10}1 & & & & & & & & & \\ 
       \textbf{Active Learning} & \cellcolor{blue!10}4 & & & & & \cellcolor{blue!10}1 & & & & & & \\ 
       \textbf{Instance Learning} & \cellcolor{blue!10}2 & & & & & & & & & & & \\ \bottomrule 
    \end{tabular}
    \caption{Evidence gap map showing which method categories have not been explored sufficiently for various task categories. Please refer to Tables~\ref{tab:taskcat} and~\ref{tab:methodex} for task and model abbreviations.}
    \label{tab:mtmap}
\end{table*}

\noindent
\textbf{Identifying under-explored and promising methods:} Annotating long tail macro-level dimensions and adaptation method categories studied by all works included in our representative sample has the additional benefit of providing a framework to identify both the most under-explored, as well as most promising methods, under various settings. Tables~\ref{tab:mtmap} and~\ref{tab:mdmap} provide evidence gap maps presenting the number of works from our sample that study the utility of various method categories on different tasks and domains respectively.\footnote{We do not include languages since our meta-analysis does not solely focus on multilingual and cross-lingual work.} The first thing we note is that both maps are highly sparse, indicating that there is little to no evidence for several combinations, many of which are worth exploring. In particular, given recent state-of-the-art advances, the following settings seem ripe for exploration:
\begin{itemize}[leftmargin=*,topsep=0pt]
\setlength\itemsep{-0.5em}
    \item \textbf{Parameter addition and freezing:} Though there are only four studies in our sample (providing positive evidence) that study parameter addition and freezing methods, we believe that given the advent of large-scale language models, these categories merit further exploration for popular task categories (TC, POS, NER, NLI, SP). Both methods attempt to improve generalization by reducing overfitting which is likely to be more prevalent with large language models, and are additionally \emph{efficient} methods that do not require a large number of extra parameters. 
    \item \textbf{Active Learning:} Studies included in our sample provide positive evidence for the use of active learning in an adaptation setting, but they have mainly evaluated on text classification (primarily sentiment analysis). We hypothesize that active learning during adaptation might also prove to be beneficial for task categories POS, NER, and SP, which require more complex, linguistically-informed annotation.
    \item \textbf{Data Selection:} Despite being similar in nature to instance weighting methods for which several studies provide positive evidence, data selection methods seem to have been under-explored. We believe that these methods might be useful for POS, NER, and SP tasks for which large-scale fortuitous data is not as easily available, and adaptation must also take into account shifts in output structure. 
\end{itemize}
Despite the scarcity of both maps, there are certain method-task and method-domain combinations for which our meta-analysis sample includes a reasonable number of studies ($>=$10\%). For these combinations, we provide a quick performance summary below:
\begin{itemize}[leftmargin=*,topsep=0pt]
\setlength\itemsep{-0.5em}
    \item \textbf{Feature Augmentation:} On text classification, 12/25 studies use FA methods as baselines. Of the remaining 13 studies, 6 provide strong positive evidence, i.e., the FA method outperforms all methods across all settings/domains tested. The remaining 7 provide mixed results, i.e., there are certain domains on which this method category doesn't work best. On semantic sequence labeling tasks like NER, 4/13 studies use FA methods as baselines, 5 show strong positive results and 4 show mixed results. Finally, on high-expertise domains, 1 study uses FA methods as baselines, 5 show strong positive results and 6 show mixed results. These observations indicate that despite their popularity, feature augmentation methods are not as strong as other method categories.
    \item \textbf{Loss Augmentation:} For text classification, 8/21 studies use LA methods as baselines. Of the remaining 13 studies, 11 provide strong positive evidence, while only 2 provide mixed results. On non-narrative domains, 9/11 studies provide strong positive evidence, while 2 provide mixed results. Based on their performance, loss augmentation methods seem to be extremely promising, especially for text classification and non-narrative domains.
    \item \textbf{Pretraining:} For text classification, 4/13 studies use pretraining as a baseline. Of the remaining 9 studies, 8 provide strong positive evidence and only one provides mixed results. Despite their relatively recent emergence, pretraining methods also seem to be extremely promising based on performance.
\end{itemize}

\begin{table}[]
    \centering
    \tiny
    \begin{tabular}{lcc}
       \toprule \backslashbox{\textbf{M}}{\textbf{D}} & \textbf{HE} & \textbf{NN} \\ \midrule
       \textbf{Feature Augmentation} & \cellcolor{blue!35}12 & \cellcolor{blue!20}8 \\ 
       \textbf{Feature Generalization} & \cellcolor{blue!10}1 & \\ 
       \textbf{Loss Augmentation} & \cellcolor{blue!20}9 & \cellcolor{blue!35}11 \\ 
       \textbf{Parameter Initialization} & \cellcolor{blue!10}1 & \cellcolor{blue!10}4\\ 
       \textbf{Parameter Addition} & & \cellcolor{blue!10}1\\  
       \textbf{Parameter Freezing} & \cellcolor{blue!10}1 & \cellcolor{blue!10}1 \\  
       \textbf{Ensemble} & \cellcolor{blue!10}2 & \cellcolor{blue!10}1 \\  
       \textbf{Instance Weighting} & \cellcolor{blue!10}2 & \cellcolor{blue!10}2 \\  
       \textbf{Data Selection} & \cellcolor{blue!10}1 & \\  
       \textbf{Pretraining} & \cellcolor{blue!10}5 & \cellcolor{blue!20}6 \\ 
       \textbf{Pseudo-Labeling} & \cellcolor{blue!10}4 & \cellcolor{blue!10}3 \\  
       \textbf{Noising/Denoising} & \cellcolor{blue!10}1 & \\  
       \textbf{Active Learning} & & \\  
       \textbf{Instance Learning} & & \\  \bottomrule
    \end{tabular}
    \caption{Evidence gap map showing indicating which method categories have not been explored sufficiently for various long tail domain categories. Note that HE and NN refer to high-expertise and non-narrative domains. Please refer to Table~\ref{tab:methodex} for model abbreviations.}
    \label{tab:mdmap}
\end{table}

\section{Which Methodological Gaps Have Greatest Negative Impact On Long Tail Performance?}
The final goal of our meta-analysis is to identify methodological gaps in developing adaptation methods for long tail domains, which provide avenues for future research. Our observations highlight three areas: (i) combining adaptation methods, (ii) incorporating external knowledge, and (iii) application to data-scarce settings.

\subsection{Combining adaptation methods}
The potential of combining multiple adaptation methods has not been systematically and extensively studied. Combining methods may be useful in two scenarios. The first one is when source and target domains differ along multiple dimensions (e.g., topic, language etc.) and different methods are known to work well for each. The second one is when methods focus on resolving issues in specific portions of the model such as feature space misalignment, task level differences etc. Combining model-centric adaptation methods\footnote{as per our categorization presented in~\S\ref{ssec:methodcat}} that tackle each issue separately may improve performance over individual approaches. Despite its utility, method combination has only been systematically explored by one meta-study from 2010. On the other hand, 23 studies apply a particular combination of methods to their tasks/domains, but do not analyze when these combinations do/do not work. We summarize both sources of evidence and highlight open questions.

\noindent
\textbf{Method combination meta-study:} \newcite{chang-etal-2010-necessity} observe that most adaptation methods either tackle shift in feature space ($P(X)$) or shift in how features are linked to labels ($P(Y | X)$). They call the former category ``unlabeled adaptation methods'' since feature space alignment can be done using unlabeled data alone. Methods from the latter category require some labeled target data and are called ``labeled adaptation methods''.\footnote{These categories do not map cleanly to our hierarchy.} Through theoretical analysis, simulated experiments and experiments with real-world data on two tasks (named entity recognition and preposition sense disambiguation), they observe: (i) combination generally improves performance, (ii) combining best-performing individual methods may not provide best combination performance, and (iii) simpler labeled adaptation algorithms (e.g., jointly training on source and target data) interface better with strong unlabeled adaptation algorithms. 

\begin{table}[t]
    \centering
    \small
    \begin{tabular}{llcc}
    \toprule \textbf{Study} & \textbf{Method} & \textbf{LT}\\ \midrule
    \multicolumn{3}{c}{\textbf{Different Coarse Categories}} \\ \midrule
    \cite{jeong-etal-2009-semi} & IW+PL & \ding{52} \\ 
    \cite{hangya-etal-2018-two} & PT+FA & \ding{52} \\ 
    \cite{cer-etal-2018-universal} & PT+LA & \ding{52} \\
    \cite{dereli-saraclar-2019-convolutional} & FA+PT & \ding{52} \\
    \cite{ji-etal-2015-closing} & FG+IW & \\
    \cite{huang-etal-2019-improving} & PI+PL & \\
    \cite{li-etal-2012-cross} & LA+PL+IW & \\
    \cite{chan-ng-2007-domain} & AL+PI+IW & \\
    \cite{nguyen-etal-2014-robust} & PL+EN & \\
    \cite{yu-kubler-2011-filling} & PL+IW & \\
    \cite{scheible-schutze-2013-sentiment} & FA+PL+DS & \\
    \cite{tan-cheng-2009-improving} & FA+IW & \\
    \cite{mohit-etal-2012-recall} & LA+PL & \\
    \cite{rai-etal-2010-domain} & AL+LA & \\
    \cite{wu-etal-2017-active} & AL+LA & \\ \midrule
    \multicolumn{3}{c}{\textbf{Same Coarse Categories}} \\ \midrule
    \cite{lin-lu-2018-neural} & PA+FA & \ding{52} \\ 
    \cite{zhang-etal-2017-aspect} & FA+LA & \ding{52} \\ 
    \cite{yan-etal-2020-multi} & FA+LA & \\ 
    \cite{yang-etal-2017-semi} & LA+PL+FA & \\ 
    \cite{gong-etal-2016-modeling} & LA+PI & \\ \midrule
    \multicolumn{3}{c}{\textbf{Same Fine Categories}} \\ \midrule
    \cite{alam-etal-2018-domain} & LA+LA & \ding{52}\\
    \cite{lee-etal-2020-pushing} & PL+PL & \\
    \cite{kim-etal-2017-cross} & LA+LA & \\
    \bottomrule
    \end{tabular}
    \caption{\small{Category combinations explored by studies that combine multiple methods. LT indicates whether long tail domains were evaluated on. Method categories explored include feature augmentation (FA), feature generalization (FG), loss augmentation (LA), parameter initialization (PI), ensembling (EN), pseudo-labeling (PL), pretraining (PT), active learning (AL), instance weighting (IW), and data selection (DS).}}
    \label{tab:combmethod}
\end{table}


\noindent
\textbf{Applying particular combinations:} Table~\ref{tab:combmethod} lists all studies that apply method combinations and fine-grained category labels from our hierarchy for the methods used. Combining methods from different coarse categories is the most popular strategy, employed by 15 out of 23 studies. 5 studies combine methods from the same coarse category, but different fine categories. They combine model-centric methods that edit different parts of the model (e.g. a feature-centric and a loss-centric method). The last 3 studies combine methods from the same fine category. Only 7 studies evaluate on at least one long tail domain.

Several studies observe performance improvements \cite{yu-kubler-2011-filling,mohit-etal-2012-recall,scheible-schutze-2013-sentiment,kim-etal-2017-cross,yang-etal-2017-semi,alam-etal-2018-domain}, mirroring the observation by \newcite{chang-etal-2010-necessity} that method combination helps. However, this is not consistent across all studies. For example, \newcite{jochim-schutze-2014-improving} find that combining marginalized stacked denoising autoencoders (mSDA) \cite{chen2012marginalized} and frustratingly easy domain adaptation (FEDA) \cite{daume-iii-2007-frustratingly} performs worse than individual methods in preliminary experiments on citation polarity classification. Both methods are feature-centric, though mSDA is a generalization method (FG) while FEDA is an augmentation method (FA). Additionally, mSDA is an unlabeled adaptation method while FEDA is a labeled adaptation method. Owing to negative results, \newcite{jochim-schutze-2014-improving} do not experiment further to find a combination that might have worked. \newcite{wright-augenstein-2020-transformer} show that combining adversarial domain adaptation (ADA) \cite{ganin2015unsupervised} with pretraining does not improve performance, but combining mixture of experts (MoE) with pretraining does. This indicates that methods from the same coarse category (model-centric) may react differently in combination settings. Similarly, studies achieving positive results do not analyze which properties of chosen methods allow them to combine well, and whether this success extends to other methods with similar properties. 

\noindent
\textbf{Open questions:} To understand method combination, we must examine the following questions:
\begin{itemize}[leftmargin=*,topsep=0pt]
\setlength\itemsep{-0.5em}
    \item Is it possible to draw general conclusions about the potential of combining methods from various coarse or fine categories?
    \item Which properties of adaptation methods are indicative of their ability to interface well with other methods? 
    \item Do task and/or domain of interest influence the abilities of methods to combine successfully?
\end{itemize}

\subsection{Incorporating external knowledge}
Most methods leverage labeled/unlabeled text to learn generalizable representations. However, knowledge from sources beyond text such as ontologies, human understanding of domain/task variation, etc., can also improve adaptation performance. This is especially true for domains with expert-curated ontologies (e.g., UMLS for biomedical/clinical text \cite{Bodenreider04theunified}). From our study sample, we observe some exploration of the following knowledge sources:

\noindent
\textbf{Ontological knowledge:} \newcite{romanov-shivade-2018-lessons} employ UMLS for clinical natural language inference via two techniques: (i) retrofitting word vectors as per UMLS \cite{faruqui-etal-2015-retrofitting}, and (ii) using UMLS concept distance-based attention. Retrofitting hurts performance, while concept distance provides modest improvements.

\noindent
\textbf{Domain Variation:} \cite{arnold-etal-2008-exploiting} and \cite{yang-eisenstein-2015-unsupervised} incorporate human understanding of domain variation in discrete and continuous feature spaces respectively, with some success (table~\ref{tab:longtailmethods}). Structural correspondence learning \cite{blitzer-etal-2006-domain} relies on manually defined pivot features common to source and target domains, and shows performance improvements.

\noindent
\textbf{Task Variation:} \cite{zarrella-marsh-2016-mitre} incorporate human understanding of knowledge required for stance detection to define an auxiliary hashtag prediction task, which improves target task performance.

\noindent
\textbf{Manual Adaptation:} \cite{chiticariu-etal-2010-domain} manually customize rule-based NER models, matching scores achieved by supervised models. 

Another source that is not explored by studies in our sample, but has gained popularity is providing task descriptions for sample-efficient transfer learning \cite{schick-schutze-2021-exploiting}. Despite initial explorations, the potential of external knowledge sources is largely under-explored.

\noindent
\textbf{Open questions:} Given varying availability of knowledge sources across tasks/domains, comparing their performance across domains may be impractical. But studies experimenting with a specific source can still probe the following questions:  
\begin{itemize}[leftmargin=*,topsep=0pt]
\setlength\itemsep{-0.5em}
    \item Can reliance on labeled/unlabeled data be reduced while maintaining the same performance?
    \item Does incorporating the knowledge source improve interpretability of the adaptation method?
    \item Can we preemptively identify a subset of samples which may benefit from the knowledge?
\end{itemize}

\subsection{Application to data-scarce settings}
\S\ref{sec:lttrends} shows that most studies test methods in a supervised setting in which labeled and/or unlabeled data is available from both source and target domains. But availability of labeled or unlabeled data is often limited for long tail domains and languages. Hence, methods should also be developed for and applied to settings that reflect real-world criteria like data availability. Data-scarce adaptation settings might be harder, but are extremely important since they closely resemble contexts in which transfer learning is likely to be used. In particular, more evaluation should be carried out in the following data-scarce settings:

\noindent
\textbf{Unsupervised Adaptation:} No labeled target data is available. Methods can use unlabeled target data or obtain distantly supervised target data from auxiliary resources (e.g., gazetteers) and user-generated signals (e.g., likes, shares, etc.). 

\noindent
\textbf{Multi-source Adaptation:} Instead of a single large-scale source dataset, smaller datasets from several source domains are available.

\noindent
\textbf{Online Adaptation:} Especially pertinent for productionizing models, in this setting, adaptation methods must learn to adapt to new domains on-the-fly. Often information about the target domain beyond the current sample may not be available. 

\noindent
\textbf{Source-free Adaptation:} A trained model must be adapted to a target domain without source domain data, either labeled or unlabeled. This setting is especially useful for domains that have strong data-sharing restrictions such as clinical data. 

Some of these settings have attracted attention in recent years. \newcite{ramponi-plank-2020-neural} comprehensively survey neural methods for unsupervised adaptation. In their survey on low-resource NLP, \newcite{hedderich2020survey} cover transfer learning techniques that reduce need for supervised target data. \newcite{wang-etal-2021-putting} list human-in-the-loop data augmentation and model updation techniques that can be used for data-scarce adaptation. However, there is room to further study application of adaptation methods in data-scarce settings. 

\noindent
\textbf{Open questions:} Broadly, two main questions in this area still remain unanswered:
\begin{itemize}[leftmargin=*,topsep=0pt]
\setlength\itemsep{-0.5em}
    \item At different levels of data scarcity (e.g., no labeled target data, no unlabeled target data, etc.), which adaptation methods perform best?
    \item Can we correlate source-target domain distance and data-reliance of adaptation methods?
\end{itemize}

\begin{table}
    \centering
    \footnotesize
    \setlength{\tabcolsep}{3.25pt}
\begin{tabular}{lccccccccc}
   \toprule & \multicolumn{3}{c}{\textbf{i2b22006}} & \multicolumn{3}{c}{\textbf{i2b22010}} & \multicolumn{3}{c}{\textbf{i2b22014}}\\ \cmidrule{2-10}
  \textbf{AM} & \textbf{P}& \textbf{R} & \textbf{F1} & \textbf{P}& \textbf{R} & \textbf{F1} & \textbf{P}& \textbf{R} & \textbf{F1} \\ \midrule
   \textbf{ZS} & 18.7 & 21.8 & 20.1 & 35.2 & 10.1 & 15.7 & 21.2 & \textbf{32.8} & 25.7 \\
   \textbf{LA} & 16.1 & 21.2 & 18.3 & 36.6 & \textbf{15.4} & \textbf{21.7} & 27.5 & 28.6 & 28.0 \\
   \textbf{PL} & \textbf{23.2} & 22.0 & \textbf{22.6} & 23.3 & 5.0 & 8.3 & \textbf{47.4} & 23.6 & \textbf{31.5} \\
   \textbf{PT} & 19.5 & \textbf{22.1} & 20.7 & \textbf{38.1} & 12.8 & 19.1 & 27.3 & 27.4 & 27.3 \\
   \textbf{IW} & 21.0 & 19.5 & 20.2 & 34.3 & 12.1 & 17.9 & 21.0 & 29.2 & 24.4 \\ \bottomrule
\end{tabular}
\caption{Performance of all adaptation methods on NER in the coarse setting. Recall that the fine adaptation method categories we evaluate are loss augmentation (LA), pseudo-labeling (PL), pretraining (PT), and instance weighting (IW).}
\label{tab:nercoarse}
\end{table}

\begin{table}
    \centering
    \footnotesize
\begin{tabular}{lcccccc}
  \toprule & \multicolumn{3}{c}{\textbf{i2b22006}} & \multicolumn{3}{c}{\textbf{i2b22014}}\\ \cmidrule{2-7}
  \textbf{AM} & \textbf{P}& \textbf{R} & \textbf{F1} & \textbf{P}& \textbf{R} & \textbf{F1} \\ \midrule
  \textbf{ZS} & 12.6 & 14.1 & 13.3 & 24.0 & \textbf{28.3} & 25.9 \\
  \textbf{LA} & 16.1 & \textbf{15.8} & \textbf{16.0} & 22.8 & 25.7 & 24.2 \\
  \textbf{PL} & \textbf{17.5} & 11.4 & 13.8 & \textbf{39.5} & 21.4 & \textbf{27.7} \\
  \textbf{PT} & 10.0 & 12.3 & 11.1 & 17.1 & 22.3 & 19.4 \\
  \textbf{IW} & 14.4 & 14.1 & 14.2 & 21.8 & 25.6 & 23.6 \\ \bottomrule
\end{tabular}
\caption{Performance of all adaptation methods on NER in the fine setting. Recall that the fine adaptation method categories we evaluate are loss augmentation (LA), pseudo-labeling (PL), pretraining (PT), and instance weighting (IW).}
\label{tab:nerfine}
\vspace{-5mm}
\end{table}

\section{Case Study: Evaluating Adaptation Methods on Clinical Narratives}
Finally, we attempt to demonstrate how our meta-analysis framework and observations can be used to systematically design case studies that can provide answers to the prevailing open questions laid out previously. As an example, we conduct a case study to evaluate the effectiveness of popularly used adaptation methods on high-expertise domains in an unsupervised adaptation setting, a burgeoning area of interest \cite{ramponi-plank-2020-neural}. Specifically, our study focuses on the question: which method categories perform best for semantic sequence labeling tasks when transferring from news to clinical narratives, given an unsupervised setting (i.e., no labeled clinical data available)? We focus on two semantic sequence labeling tasks: entity extraction and event extraction.

\subsection{Datasets}
We use the following entity extraction datasets:
\begin{itemize}[leftmargin=*,topsep=0pt]
\setlength\itemsep{-0.5em}
    \item \textbf{CoNLL 2003} \cite{tjong-kim-sang-de-meulder-2003-introduction}: News stories annotated with four entity types: persons, organizations, locations, and miscellaneous names.
    \item \textbf{i2b2 2006} \cite{uzuner2007evaluating}: Medical discharge summaries annotated with PHI (private health information) entities of eight types: patients, doctors, locations, hospitals, dates, IDs, phone numbers, and ages.
    \item \textbf{i2b2 2010} \cite{uzuner20112010}: Discharge summaries annotated with three entity types: medical problems, tests and treatments.
    \item \textbf{i2b2 2014} \cite{stubbs2015annotating}: Longitudinal medical records annotated with PHI entities of eight broad types: name, profession, location, age, date, contact, IDs, and other.
\end{itemize}
All entities are annotated in IOB format. For event extraction, we use the following datasets:
\begin{itemize}[leftmargin=*,topsep=0pt]
\setlength\itemsep{-0.5em}
    \item \textbf{TimeBank} \cite{pustejovsky2003timebank}: News articles annotated with events.
    \item \textbf{i2b2 2012} \cite{sun2013evaluating}: Discharge summaries annotated with events.
    \item \textbf{MTSamples} \cite{naik-etal-2021-adapting}: Medical records annotated with events (test-only).
\end{itemize}
CoNLL 2003 and TimeBank are the source datasets for all entity and event extraction experiments respectively, while the remaining are target datasets. We focus on English narratives only. Among the NER datasets, the label sets for i2b22006 and i2b22014 can be mapped to the label set for CoNLL2003, however the label set for i2b22010 is quite distinct and cannot be mapped. Therefore, we evaluate NER in two settings: \emph{coarse} and \emph{fine}. In the coarse setting, the model only detects entities, but does not predict entity type, whereas in the fine setting, the model detects entities and predicts types. 
\begin{table}
    \centering
    \footnotesize
\begin{tabular}{lcccccc}
   \toprule & \multicolumn{3}{c}{\textbf{i2b22012}} & \multicolumn{3}{c}{\textbf{MTSamples}}\\ \cmidrule{2-7}
  \textbf{AM} & \textbf{P}& \textbf{R} & \textbf{F1} & \textbf{P}& \textbf{R} & \textbf{F1} \\ \midrule
   \textbf{ZS} & 48.8 & 15.3 & 23.3 & 91.4 & 48.0 & 63.0 \\
   \textbf{LA} & \textbf{51.7} & \textbf{19.0} & \textbf{27.8} & 88.1 & \textbf{58.5} & \textbf{70.3} \\
   \textbf{PL} & 44.1 & 11.4 & 18.2 & \textbf{91.8} & 39.3 & 55.1 \\
   \textbf{PT} & 41.5 & 10.4 & 16.6 & 90.2 & 46.3 & 61.2 \\
   \textbf{IW} & 50.5 & 18.1 & 26.6 & 90.6 & 48.4 & 63.1 \\ \bottomrule
\end{tabular}
\caption{Performance of all adaptation methods on event extraction. Recall that the fine adaptation method categories we evaluate are loss augmentation (LA), pseudo-labeling (PL), pretraining (PT), and instance weighting (IW).}
\label{tab:events}
\vspace{-5mm}
\end{table}

\subsection{Adaptation Methods}
The baseline model for both tasks is a BERT-based sequence labeling model that computes token-level representations using BERT, followed by a linear layer that predicts entity/event labels. We compare the performance of adaptation methods from the top 5 fine categories most frequently applied to high-expertise domains as per our analysis (figure~\ref{fig:exfine}), on top of this BERT baseline. Since feature augmentation (FA) methods require some target labeled data to train target-specific weights and our focus is on an unsupervised setting, our study tests the remaining four categories:
\begin{itemize}[leftmargin=*,topsep=0pt]
\setlength\itemsep{-0.5em}
    \item \textbf{PL:} From pseudo-labeling, we test the self-training method. Self-training first trains a sequence labeling model on the source dataset (news), then uses this model to generate labels for unlabeled target data (clinical narratives). High-confidence predictions from the ``pseudo-labeled'' clinical data are combined with source data to train a new sequence labeling model. This process can be repeated iteratively. 
    \item \textbf{LA:} From loss augmentation, we test adversarial domain adaptation \cite{ganin2015unsupervised}. This method learns domain-invariant representations by adding an adversary that predicts an example's domain and subtracting the loss from this adversary from the overall model loss. This setup is trained in a two-stage alternating optimization process (complete details in \newcite{ganin2015unsupervised}). 
    \item \textbf{PT:} From pretraining, we test domain-adaptive pretraining as described by \newcite{gururangan-etal-2020-dont}. This method tries to improve target domain performance of BERT-based models by continual masked language modeling pretraining on unlabeled text from the target domain.  
    \item \textbf{IW:} From instance weighting, we test classifier-based instance weighting. This method trains a classifier on the task of predicting an example's domain, then runs the classifier on all source domain examples and uses target domain probabilities as weights. Source examples that ``look'' more like the target domain get higher weights, improving performance on the target domain. We perform interleaved training, recomputing source weights after each model training pass.
\end{itemize}

\subsection{Results}
Tables~\ref{tab:nercoarse} and~\ref{tab:nerfine} show the results of all adaptation methods on coarse and fine NER, while table~\ref{tab:events} shows results on event extraction. ZS indicates baseline model scores in a zero-shot setting, i.e., training on source and testing on target with no adaptation. From these tables, we can see that the best-performing method categories are loss augmentation and pseudo-labeling across different settings. Loss augmentation methods work best for event extraction. For coarse NER, pseudo-labeling methods work better on target datasets whose labels can be mapped to the source (i.e., \emph{closer} transfer tasks). For i2b22010, which is more distant transfer, loss augmentation works best. The effectiveness of pseudo-labeling is interesting because they often suffer from the pitfall of propagating errors made by the source-trained model, which may in part explain their poor performance on i2b22010. Early work on applying these methods to parsing showed negative results or minor improvements \cite{charniak1997statistical,steedman-etal-2003-bootstrapping}, but these methods have shown more promise in recent years with advances in embedding representations. Finally, for fine NER, loss augmentation and pseudo-labeling do better on i2b22006 and i2b22014 respectively. Pretraining is not the best-performing method in any setting, which may be a side effect of continual pretraining leading to some forgetting, negatively impacting an unsupervised setting. This highlights the need to systematically compare adaptation methods under data-scarce settings because the ranking of methods can change based on the availability and quality of domain-specific data. 

\section{Conclusion}
This work presents a qualitative meta-analysis of 100 representative papers on domain adaptation and transfer learning in NLU, with the aim of understanding performance of adaptation methods on the long tail. Through this analysis, we assess current trends and highlight methodological gaps that we consider to be major avenues for future research in transfer learning for the long tail. We observe that long tail coverage in current research is far from comprehensive, and identify two properties of adaptation methods that may improve long tail performance, but have been under-explored. Additionally, we identify three major gaps that must be addressed to improve long tail performance: (i) combining adaptation methods, (ii) incorporating external knowledge and (iii) application to data-scarce adaptation settings. Finally, we demonstrate the utility of our meta-analysis framework and observations in guiding the design of systematic meta-experiments to address prevailing open questions by conducting a systematic evaluation of popular adaptation methods for high-expertise domains in a data-scarce setting. This case study reveals interesting insights about the adaptation methods evaluated and shows that significant progress can be made towards developing a better understanding of adaptation for the long tail by conducting such experiments.

\section*{Acknowledgements}
This research was supported in part by the Intramural Research Program of the National Institutes of Health, Clinical Research Center and through an Inter-Agency Agreement with the US Social Security Administration. The views and conclusions contained herein are those of the authors and should not be interpreted as necessarily representing the official policies or endorsements, either expressed or implied, of the NIH, or the US Government. The authors would like to thank Emma Strubell, Matt Gormley, Luke Zettlemoyer, Abhilasha Ravichander and Khyathi Chandu for their feedback on early drafts of this work. The authors are also extremely grateful for the valuable reviews provided by the anonymous reviewers and the action editor, Benjamin van Durme, which significantly improved our final version. 

\bibliography{main-3647-Naik}
\bibliographystyle{acl_natbib}

\end{document}